%% file: bmvc_final.tex
\documentclass{bmvc2k}

\title{AMICO: Amodal Instance Composition}

\addauthor{Peiye Zhuang}{peiye@illinois.edu}{1}
\addauthor{Jia-Bin Huang}{jbhuang@fb.com}{23}
\addauthor{Ayush Saraf}{ayush29feb@fb.com}{3}
\addauthor{Xuejian Rong}{xrong@fb.com}{3}
\addauthor{Changil Kim}{changil@fb.com}{3}
\addauthor{Denis Demandolx}{denisd@fb.com}{3}

\addinstitution{
University of Illinois Urbana-Champaign
}
\addinstitution{
University of Maryland, College Park
}
\addinstitution{
Facebook
}

\runninghead{Zhuang et al.}{Amodal Instance Composition}

\usepackage{times}
\usepackage{epsfig}
\usepackage{graphicx}
\usepackage{amsmath}
\usepackage{amssymb}

\usepackage{booktabs}
\usepackage{float}
\usepackage{textcomp}
\usepackage{xcolor}
\usepackage{bm}
\usepackage{subfigure} 
\usepackage{paralist}
\usepackage{algorithm, algpseudocode}
\usepackage{caption}
\usepackage{multirow}

\newcommand{\topic}[1]
{
\vspace{1mm}\noindent\textbf{#1}
}

\algnewcommand{\Input}{\item[\textbf{Input:}]}%
\algnewcommand{\Output}{\item[\textbf{Output:}]}
\algnewcommand{\RETURN}{\item[\textbf{Return:}]}

\begin{document}

\maketitle

\input{01-abs}

\input{02-intro}

\input{03-related}

\input{04-method}
\input{05-exp}

\input{06-con}
\bibliography{bmvc_final}
\clearpage
\input{07-sup}

\end{document}

%% file: 01-abs.tex
\vspace{-1cm}
\input{figure_latex/01-teaser}
\vspace{-1cm}
\begin{abstract}
\noindent
Image composition aims to blend multiple objects to form a harmonized image. Existing approaches often assume precisely segmented and intact objects. Such assumptions, however, are hard to satisfy in unconstrained scenarios. We present Amodal Instance Composition for compositing imperfect---potentially incomplete and/or coarsely segmented---objects onto a target image. We first develop object shape prediction and content completion modules to synthesize the amodal contents. We then propose a neural composition model to blend the objects seamlessly. Our primary technical novelty lies in using separate foreground/background representations and blending mask prediction to alleviate segmentation errors. Our results show state-of-the-art performance on public COCOA and KINS benchmarks and attain favorable visual results across diverse scenes. We demonstrate various image composition applications such as object insertion and de-occlusion.
\vspace{-5mm}
\end{abstract}

%% file: figure_latex/01-teaser.tex
\begin{figure}[th]

\includegraphics[width=\linewidth]{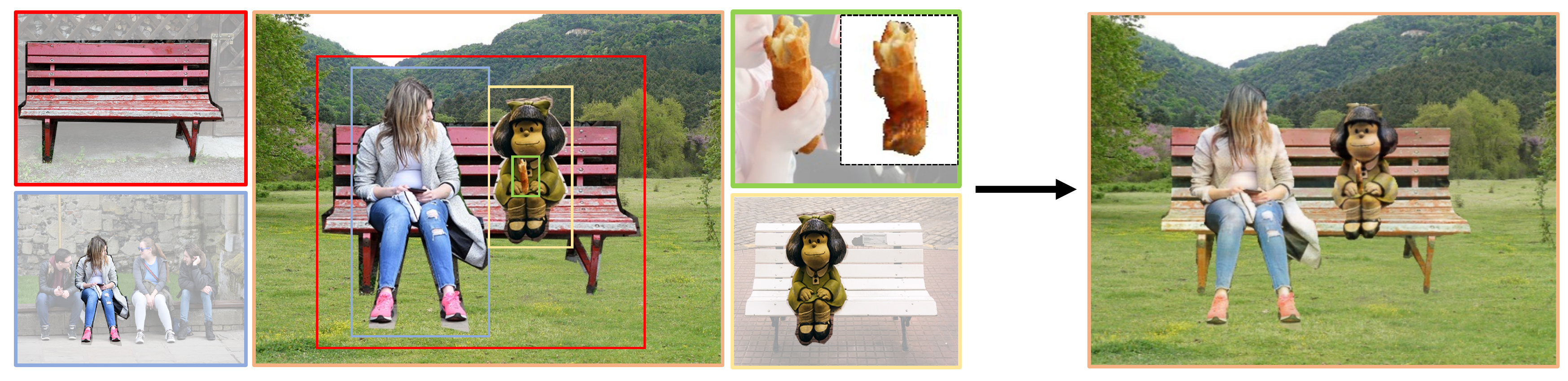}
\begin{minipage}[h]{\linewidth}
$ \qquad  \qquad \qquad\ $(a) Compositing components $\qquad \qquad \qquad $ (b) Composed image
\centering
\end{minipage}

\vspace{0.2cm}
\caption{
\textbf{Composition with imperfect instances.} Our method takes an ordered collection of imperfect instances (i.e., partially occluded and/or coarsely cropped) from multiple source images (instances identified by colored bounding boxes) and a background image as inputs \emph{(a)}, and produces harmonized composition \emph{(b)}. We achieve this via a unified framework that estimates the precise shape and content of each object and adjusts the object appearances to be mutually compatible.
}
\label{fig:teaser}
\end{figure}


%% file: 02-intro.tex
\vspace{-.5cm}
\section{Introduction}
\label{sec:intro}
\vspace{-.2cm}

Image composition is a classic photo editing task that combines color-inconsistent objects from multiple source images into one composite image.
Most existing approaches often assume intact (without occlusion) and precisely segmented instances~\cite{azadi2018compositional,lin2018st,chen2019toward,cong2020dovenet}.
Such instances, however, may be challenging to obtain in real-world scenarios due to complex object shapes and occlusions in an image, e.g., bread in Figure~\ref{fig:teaser}(a). 

We introduce the Amodal Instance Composition problem: compositing \emph{imperfect} object instances (e.g., coarsely segmented or with incomplete shape/appearance due to occlusion) onto a target background image.
The amodal instance composition poses several novel challenges for conventional image composition problems.
First, given an object instance under occlusion, we need to perform amodal segmentation, estimating the object's full spatial extent beyond its visible regions, and then complete (hallucinate) the occluded regions of the selected object instance. Prior amodal instance completion methods~\cite{zhan2020self, ling2020variational} directly applied classic image inpainting approach ~\cite{liu2018image} for object completion. However, amodal object completion is different from image inpainting due to complex occlusion relationships, object shapes, and materials~\cite{carreira2011cpmc, arbelaez2012semantic, xie2020segmenting}. As a result, the prior methods~\cite{zhan2020self, ling2020variational} often produce unrealistic results for object content completion.



Second, we need to adjust the appearance of the (completed) object instances to make them compatible with the background. When existing amodal instance completion methods~\cite{zhan2020self, ling2020variational} demonstrate image manipulation tasks, e.g., to insert (completed) amodal instances to a new background, we observe that these methods take no consideration for color consistency. Unavoidably, they result in unrealistic composite images when the instances have distinct colors from the background. 
Moreover, the current composition methods \cite{azadi2018compositional,lin2018st,chen2019toward,cong2020dovenet} often produce visible artifacts when imprecise masks are used as input.

In this paper, we present a fully automatic system to tackle the Amodal Instance Composition problem. 
Our method consists of three main modules tailored explicitly for addressing the above challenges.
1) \emph{Object content completion}: Our object content completion module uses amodal and visible masks to synthesize the appearance of the missing regions. Our critical insight here is to use visible object regions only (instead of using the entire image as input). 
2) \emph{Image composition}: In contrast to prior harmonization methods that use a single image (with object instance copy-and-pasted onto the background) as input, our model takes a \emph{separate} background image and object instance as inputs and produces RGBA (color and opacity) layers describing the appearance-adjusted object instance. 
3) \emph{Amodal mask prediction}: An amodal mask prediction module is trained offline and applied during inference to detect and recognize the occluded region of a given object. The estimated missing region will then be completed by our object content completion module.
We validate that our proposed design leads to favorable performance on the publicly available COCOA and KINS datasets. 

\topic{We summarize our main contributions as follows.} 
\begin{compactitem}
\item We introduce the amodal instance composition task and present a learning-based system and the corresponding training strategies.
\item We show favorable results against existing approaches on representative benchmarks and demonstrate various practical applications of amodal image composition. 

\end{compactitem}

%% file: 03-related.tex
\input{figure_latex/02-model}
\vspace{-0.4cm}
\section{Related work}
\label{sec:rel}
\vspace{-0.3cm}

\topic{Image composition} seeks to compose and blend objects from multiple source images such that the new composite image appears photorealistic by harmonizing the colors of the foreground instances. 
Earlier work uses transparency maps~\cite{porter1984compositing} or performs linear blending over multiple frequency bands~\cite{burt1987laplacian,brown2003recognising}.
These methods, however, do not handle scenarios where the appearances of the object instances are not \emph{compatible} with the background.
To harmonize the composites, previous methods use color matching techniques such as applying color gradient-domain compositing~\cite{perez2003poisson,levin2004seamless,tao2013error} and statistical features~\cite{reinhard2001color,Efros2007color,xue2012understanding}.
Data-driven approaches (e.g.,~\cite{johnson2010cg2real}) retrieve images with similar layouts from a large-scale database for compositing.
Recent learning-based image composition methods demonstrate favorable performance~\cite{tsai2017deep,cong2020dovenet,zanfir2020human,sofiiuk2020foreground,zhang2020deepblending}.
Our work also focuses on learning-based color-consistent harmonization. 
Unlike existing work that uses a single composed image as input, we show that using \emph{layered inputs} (e.g., separate foreground/background images) helps boost the harmonization quality for imperfect inputs. 
 
The mask refinement step in our proposed composition module is also relevant to \textbf{image matting}. 
Concretely, the image matting task estimates an accurate alpha matte that separates the foreground instance from the background given manually created trimaps by users~\cite{sun2004poisson, levin2007closed, levin2008spectral, aksoy2017designing, xu2017deep}. 
In contrast, our proposed approach takes as input imperfect instances and a background image (i.e., no trimaps) and produces RGBA layers with the aim of photorealistic composition.

Another line of research focuses on automatically placing an instance into a target background in a \emph{geometrically consistent} manner by applying geometric transformations on the selected instance~\cite{lin2018st, zhan2019adaptive, zhang2020placement}. 
Our work focuses on color consistency instead.

\topic{Image completion} focuses on synthesizing missing contents within one target image. 
Early methods search the good matching patches from valid image regions~\cite{efros2001image, criminisi2004patchworks,barnes2009patchmatch,huang2014image} or large-scale datasets~\cite{hays2007scene} to complete the holes. 
Recent deep learning-based approaches~\cite{pathak2016context, yang2017high, iizuka2017globally, liu2018image, nazeri2019edgeconnect, yu2019free, zheng2019pluralistic, zeng2020high, yi2020contextual} build upon the Context Encoder approach~\cite{pathak2016context}, which extends early CNN-based inpainting to large masks using Generative Adversarial Networks~\cite{goodfellow2014generative}. 
Among which, partial convolution ~\cite{liu2018image} and gated convolution~\cite{yu2019free} are proposed to address the issues of visual artifacts using vanilla convolutions. 
Further, auxiliary semantic information is leveraged to enhance inpainting performance such as edges~\cite{nazeri2019edgeconnect}, segmentation~\cite{song2018spg, liao2020guidance} and foreground object contours~\cite{xiong2019foreground}. 
In ~\cite{zhan2020self, ling2020variational}, a partial convolution network~\cite{liu2018image} was employed to synthesize the \emph{appearance} of the occluded content given predicted object amodal masks (the task also defined as \textbf{amodal instance completion}). 
Consequently, we compare our object content completion network with several representative image inpainting methods~\cite{liu2018image, nazeri2019edgeconnect, zeng2020high, yi2020contextual} and amodal instance completion methods~\cite{zhan2020self, ling2020variational} in Section~\ref{sec:exp}.

%% file: figure_latex/02-model.tex
\begin{figure}[th]
\centering
\includegraphics[width=\linewidth]{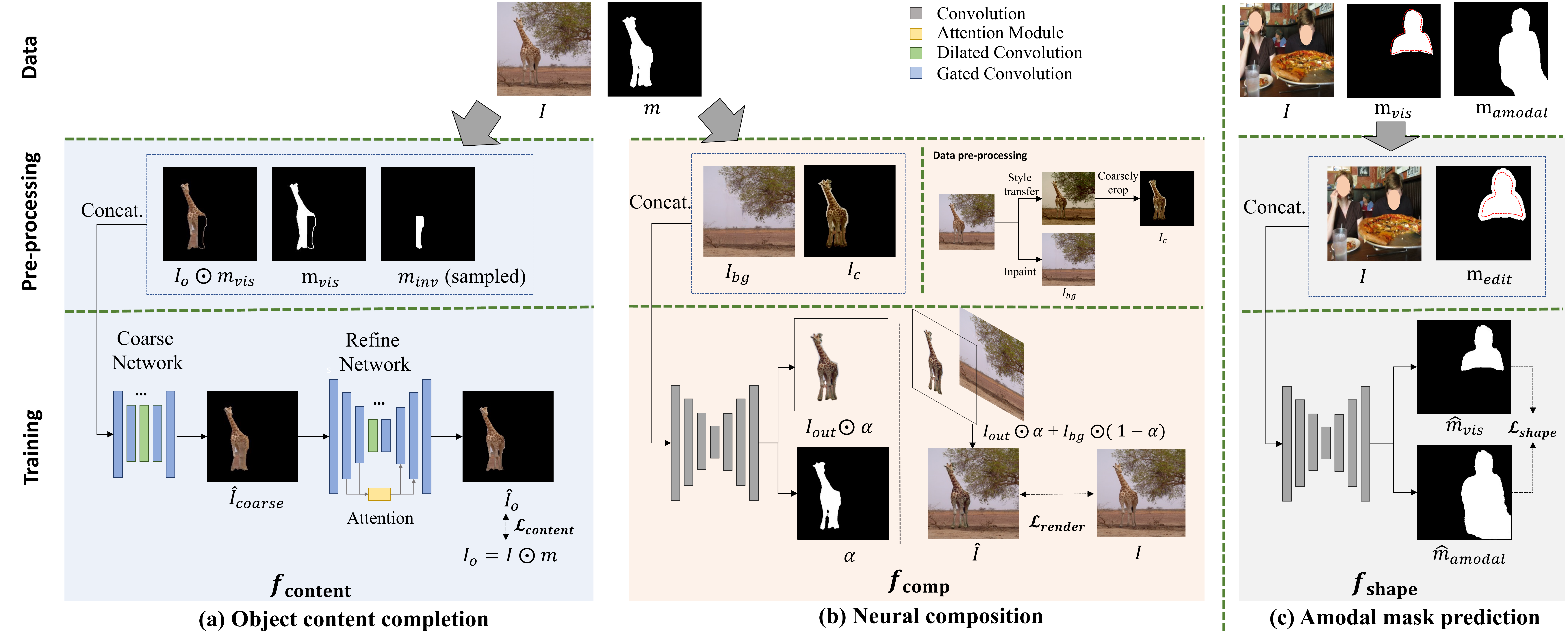}
\vspace{0.1cm}
\caption{
\textbf{Overall framework composed of three modules.} (a) An object content completion net $f_\text{content}$ targets to generate missing regions for an object given triplet images as input, including a masked object $I_o \odot m_{vis}$ and two mutually exclusive binary masks $m_{vis}$ and $m_{inv}$ marking the \emph{visible} and \emph{invisible} region of the object (``$\odot$'' denotes element-wise product). The generator in $f_\text{content}$  has two stages and a discriminator (not shown due to limited space) is used for training the generator. 
(b) A neural composition net $f_\text{comp}$ takes as input a background image $I_{bg}$ and edited object $I_c$ (refer to Sections~\ref{sec:method}-\ref{sec:exp} for details) and predicts RGBA layers, $I_{out}$ and $\alpha$, for the object. With the estimated alpha map $\alpha$, we obtain the combined image, $\hat{I}$, via alpha-blending.
(c) An amodal mask prediction net $f_\text{shape}$ (trained offline) takes as input an image $I$ and a mask $m_{edit}$ that marks the visible region of an object in $I$.
}

\vspace{-.5cm}
\label{fig:model}
\end{figure}


%% file: 04-method.tex
\input{figure_latex/completion}
\vspace{-.5cm}
\section{Our Method}
\label{sec:method}
\vspace{-0.3cm}
\noindent
We address the problem of composing multiple input images with undesirable properties, such as, those that are incomplete, coarsely cropped, occluded, or having inconsistent colors. 
For this, we introduce an object content completion network $f_\text{content}$ (Section \ref{sec:completion}), a neural compositing network $f_\text{comp}$ (Section \ref{sec:renderer}), and an amodal mask prediction network $f_\text{shape}$ (Section \ref{sec:amodal}). 
We show the overall framework, training strategies and inference procedures in Figure~\ref{fig:model} and describe them accordingly.
 
To begin with, we denote an instance segmentation dataset as $\mathcal{D} = \{(I^{(i)}, m^{(i)}, m^{(i)}_{amodal})\}_{i=1}^N$, where $I^{(i)} \in \mathbb{R}^{H\times W \times 3}$ refers to an image, and 
$m^{(i)} \text{ and } m^{(i)}_{amodal} \in \{0, 1\}^{H\times W \times 1}$ are two binary masks that mark the visible and the intact (i.e., both the visible and the invisible) region of an object in $I^{(i)}$, respectively. $N$ is the size of the dataset. 
To avoid clutter, we use simplified notations, $I$, $m$, and $m_{amodal}$ in the following sections.

\vspace{-0.3cm}
\subsection{Object content completion network}
\label{sec:completion}

\noindent
One critical step to compose occluded objects into a new background image is to complete the appearance of the invisible regions. 
Since there are no available paired images of the same object with and without occlusion, we manually construct our own paired data to train an object completion network, $f_\text{content}$, in a self-supervised manner. 

Concretely, we mask out part of an object and train $f_\text{content}$ to recover the missing content.
As shown in Figure~\ref{fig:model} \textit{(a)}, we separate the object mask $m$ into two mutually exclusive parts: 1) a visible mask $m_{vis}$, and 2) an invisible mask $m_{inv}$, where $m = m_{vis} \cup m_{inv}$.
The input of $f_\text{content}$ is a triplet consisting of a masked object image $I_o \odot m_{vis}$, and the two binary masks $m_{vis}$ and $m_{inv}$, where $I_o$ refers to the instance image, defined as  $I_o = I \odot m$. 
To obtain $m_{inv}$, we randomly sample a mask from dataset $\mathcal{D}$ that has overlapped with $m$. 
The overlapped region is captured by $m_{inv}$ with the overlapping ratio $\frac{\Vert m_{inv} \Vert_1}{\Vert m \Vert_1} \in [0.1, 0.9]$. 

Our object completion network $f_\text{content}$ consists of a two-state generator $G$ and a discriminator $D$, inspired by image inpainting techniques~\cite{yu2018generative, yu2019free,yi2020contextual}.
We use a discriminator $D$ to distinguish the input image source as either real or synthetic. 
In the generator $G$, a coarse network produces a rough completion result, notated as $\hat{I}_{coarse}$, and a refine network generates a finer completed image, namely $\hat{I}_o$. 
More detailed model architecture is presented in the supplementary.
Therefore, we train the generator $G$ with the loss function $\mathcal{L}_\text{content}$:
\vspace{-0.15cm}
\begin{equation}
 \mathcal{L}_\text{content} = 
 \lambda_1 \mathcal{L}_{\text{refine}} +  \lambda_2 \mathcal{L}_{\text{coarse}}  + \lambda_3 \mathcal{L}_{\text{adv}},
\end{equation}
where $\lambda_1$, $\lambda_2$, and $\lambda_3$ are coefficients of the loss terms.

Firstly, $\mathcal{L}_{\text{refine}}$ is the object reconstruction loss for $\hat{I}_o$:
\vspace{-0.15cm}
\begin{equation}
    \mathcal{L}_{\text{refine}} = \quad \mathbb{E}_{(I_o,m) \sim \mathcal{D}, \hat{I}_o \sim \mathcal{D}'|(I_o, m)} [
    \Vert \hat{I}_o- I_o \Vert_1 \odot m_{inv}\  + \beta \Vert \hat{I}_o - I_o \Vert_1 \odot m_{vis} ].
\vspace{-0.15cm}
\end{equation}
where, $\beta$ = 5. We refer to the distribution of $\hat{I}_o$ via $\mathcal{D}'$ conditioned on $(I, m)$, formally written as $\mathcal{D}'|(I, m)$. $\mathcal{L}_{\text{coarse}}$ is the same reconstruction loss except computing with $\hat{I}_{coarse}$.

We use the WGAN-GP \cite{gulrajani2017improved} loss to update both $G$ (via $\mathcal{L}_{\text{adv}}$) and $D$ (via $\mathcal{L}_\text{d}$):
\vspace{-0.15cm}
 \begin{equation}
\mathcal{L}_{\text{adv}} = - \mathbb{E}_{\hat{I}_o \sim \mathcal{D}'| (I_o, m)} [D(\hat{I}_o)],
\end{equation}
\vspace{-.7cm}
\begin{equation}
\centering
 \mathcal{L}_\text{d} = \, \mathbb{E}_{\hat{I}_o \sim \mathcal{D}'| (I_o, m)} [D(\hat{I}_o)]\ - \mathbb{E}_{I_o \sim \mathcal{D}}[D(I_o)] \ + 
\sigma_1 \mathbb{E}_{\hat{I}_o \sim \mathcal{D}'| (I_o, m)} [ \ \Vert \nabla_{\hat{I}_o} D(\hat{I}_o) \Vert_2 - 1\ ]^2,
\end{equation}
\vspace{-0.15cm}
where $\sigma_1 = 10$ is a weight for the gradient penalty term. $D$ and $G$ are trained alternatively. 

\input{figure_latex/comp_gt}
\vspace{-0.3cm}
\subsection{Neural compositing network}
\label{sec:renderer}
\vspace{-0.1cm}
\noindent
We propose a neural compositing network, $f_\text{comp}$, shown in Figure~\ref{fig:model} \textit{(b)}, to blend multiple objects into a single coherent image. 
The composition network $f_\text{comp}$ should be robust to objects that could be imperfectly cropped and have inconsistent appearances with the background image. 
For this, $f_\text{comp}$ takes a background image $I_{bg}$ and an edited object $I_c$ as input and generates RGBA layers, including $I_{out}$ and $\alpha$, for the object. 
With the output layers, we obtain a reconstructed image $\hat{I}$ through standard alpha blending.
In this case, $I_{out}$ contains the color-transferred object with its appearance close to the background and an $\alpha$ map helps refine the object shape. 
We introduce the implementation details in following subsections.

\topic{Data preparation.}
We train $f_\text{comp}$ in a self-supervised manner. For this, we employ two off-the-shelf modules: 1) an \textit{image inpainting module}~\cite{yi2020contextual}, denoted by Inpainting($\cdot$), for background completion; and 2) a \textit{color transfer module}~\cite{yoo2019photorealistic}, denoted by ColorTransfer($\cdot$), for foreground color modification. Formally, we have
\vspace{-0.1cm}
\begin{equation}
    I_{bg} = \text{Inpaint}(I \odot (1-m), m), 
    I_c = \text{ColorTransfer}(I, I_{ref}) \odot \text{Dilate}(m, iter). 
\end{equation}
\vspace{-0.1cm}
Inpainting($\cdot$) completes the masked background region marked in $m$, and the ColorTransfer($\cdot$) transfers colors from a randomly sampled reference image $I_{ref} \in \mathcal{D}$ to the target $I$. Dilate($\cdot$) simulates the coarsely cropping step by randomly enlarging the cropped region for the object with $iter\sim\mathcal{U}\{0, \cdots, 25\}$ pixels.

\topic{Model optimization}. As we obtain the compositing result $\hat{I}$ via alpha blending:
\vspace{-0.1cm}
\begin{equation*}
    \hat{I} = I_{out} \odot \alpha + I_{bg} \odot (1 - \alpha),
\end{equation*}
\vspace{-0.1cm}
we use three loss terms to optimize $f_\text{comp}$: a reconstruction loss $\mathcal{L_\text{recon}}$, a mask loss $\mathcal{L_\text{mask}} $, and a regularization loss  $\mathcal{L_\text{reg}}$ for $\alpha$.

First, $\mathcal{L_\text{recon}}$ assesses how well the neural compositing model reconstructs $I$. We express it via $\ell_1$ loss:
\begin{equation}
    \mathcal{L_\text{recon}} = \mathbb{E}_{(I, m) \sim \mathcal{D}} \Vert \hat{I}- I\Vert_1.
\end{equation}

Second, a mask loss $\mathcal{L_\text{mask}}$ on the $\alpha$ layer is used to encourage the learned $\alpha$ layer to match the exact object segment. Formally, we have
\vspace{-0.1cm}
\begin{equation}
    \mathcal{L_\text{mask}} = \mathbb{E}_{(I\!, m) \sim\mathcal{D}} 
    \frac{\Vert m \odot (1\!-\!\alpha) \Vert_1 }{2\Vert m\Vert_1} \! + \!
    \frac{\Vert (1\!-\!m) \odot \alpha \Vert_1 }{2\Vert 1-m \Vert_1}.
\end{equation}

Last, inspired by~\cite{lu2020retiming},  we apply a regularization loss, $\mathcal{L_\text{reg}}$, consisting of an $L_1$ norm and an approximation $L_0$ norm, to encourage $\alpha$ to be spatially sparse:
\begin{equation}
    \mathcal{L_\text{reg}} = \mathbb{E}_{(I, m) \sim\mathcal{D}} \gamma \Vert \alpha \Vert_1 + 2 \cdot \text{Sigmoid} (5 \cdot \alpha) -1 ,
\end{equation}
where $\gamma$ controls the relative weight ratio between the two terms.

Hence, the total loss of the neural compositing model is 
\vspace{-0.1cm}
\begin{equation}
    \mathcal{L}_\text{comp} = \mathcal{L_\text{recon}} + \lambda_4 \mathcal{L_\text{mask}} + \lambda_5 \mathcal{L_\text{reg}},
\end{equation}

where $\lambda_4$ and $\lambda_5$ are loss weights. We approximate all aforementioned expectations by empirical sampling.
\input{figure_latex/05-amo-comp}

\vspace{-0.2cm}
\subsection{Amodal mask prediction network}
\label{sec:amodal}
\vspace{-0.1cm}
We employ an amodal mask prediction network, defined as $f_\text{shape}$, that is trained offline and applied during inference to predict the visible and the intact regions of an object, defined as $\hat{m}_{vis}$ and $\hat{m}_{amodal}$, respectively. The estimated missing region is then completed by the content completion network $f_\text{content}$. As shown in Figure~\ref{fig:model} \textit{(c)}, the input of the amodal mask prediction network, $f_\text{shape}$, is an image $I$ and a binary $m_{edit}$ that roughly indicates the visible region. We obtain $m_{edit}$ from $m_{vis}$ via editing (e.g., dilation and erosion) to increase the robustness of $f_\text{shape}$ for imperfect inference cases where the mask of the visible area may not be accurate. 
We formulate the loss function for optimizing $f_\text{shape}$ as follows:
\vspace{-0.1cm}
\begin{equation}
    \mathcal{L}_{\text{shape}} = \mathbb{E}_{\mathcal{D}}\ \ \ell (\hat{m}_{vis}, m_{vis}) +  \ell (\hat{m}_{amodal}, m_{amodal}),
\end{equation}
\vspace{-0.1cm}
where $\ell$ is a weighted binary cross-entropy (BCE) loss computed on both regions inside and outside of the object area:
\begin{equation}
     \ell(m_1, m_2) = \omega \cdot \text{BCE}(m_1 \odot m_2, m_2) + \text{BCE}((1-m_1) \odot (1-m_2), 1-m_2),
\end{equation}
\vspace{-0.1cm}
where $\omega=5$ and $\odot$ represents element-wise product.

\vspace{-0.1cm}
\subsection{Model inference}
\label{sec:model_infer}

\noindent
Given a background image $I_{bg}$, and object images $X = \{x^{(1)}, x^{(2)}, ..., x^{(M)}\}$, where $x^{(j)} = (I^{(j)}, m_{vis}^{(j)})$, $j \in \{1, 2,...,M\}$ and $M= |X|$. Note that $m_{vis}^{(j)}$ is not required to be precise during inference. For occluded objects (if any), $f_\text{shape}$ and $f_\text{content}$ are applied in sequence to predict and synthesize the invisible regions. Afterwards, the objects are composed with the background image iteratively by $f_\text{comp}$, and the background image is progressively updated with each object composition. More detailed algorithms are presented in the supplementary.

%% file: figure_latex/completion.tex
\begin{figure}[t]
\centering
\includegraphics[width=\linewidth]{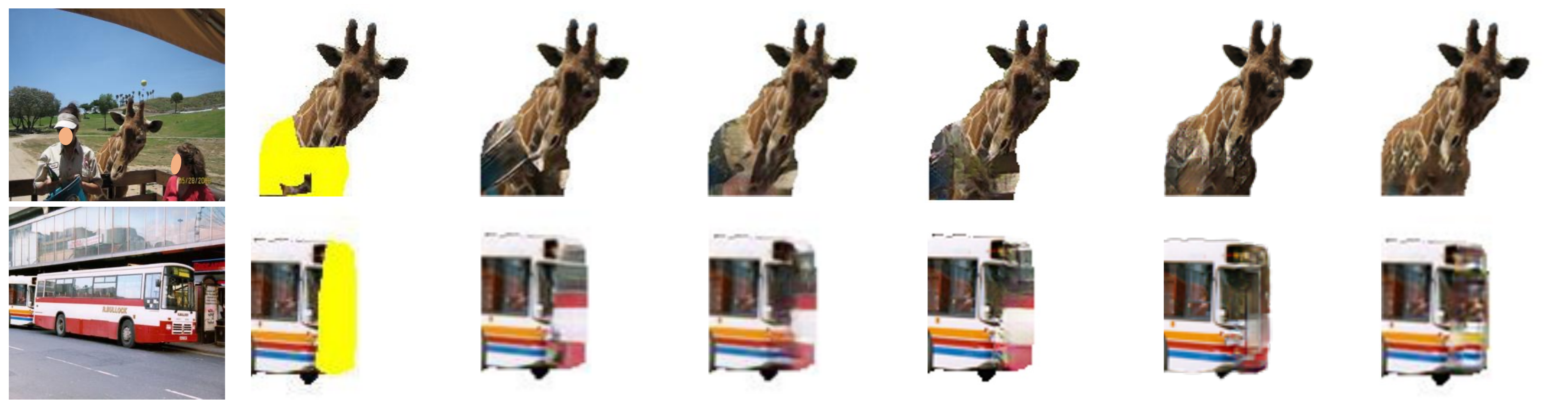}
\begin{minipage}{\linewidth}
Original \hspace{0.3cm} Occ. object \hspace{0.3cm} 
GConv\cite{yu2019free} \hspace{0.3cm} 
CRA\cite{yi2020contextual}  \hspace{0.3cm}
EdgeCon.\cite{nazeri2019edgeconnect} \hspace{0.3cm}
De-occ.\cite{zhan2020self}\hspace{0.3cm} Ours
\end{minipage}
\vspace{0.2cm}
\caption{
\textbf{Samples of object completion for occluded object instances.} 
From left to right: original images from COCO test dataset with available visible and amodal masks annotated by ~\cite{zhu2017semantic} (\textit{Column 1}); objects with occluded region in yellow and background region in white (\textit{Column 2}); results from the baseline methods and ours (\textit{Columns 3--7}) .}
    \label{fig:amodal_completion}
\vspace{-.5cm}
\end{figure}

%% file: figure_latex/comp_gt.tex
\begin{figure*}[th]
\centering
\includegraphics[width=\linewidth]{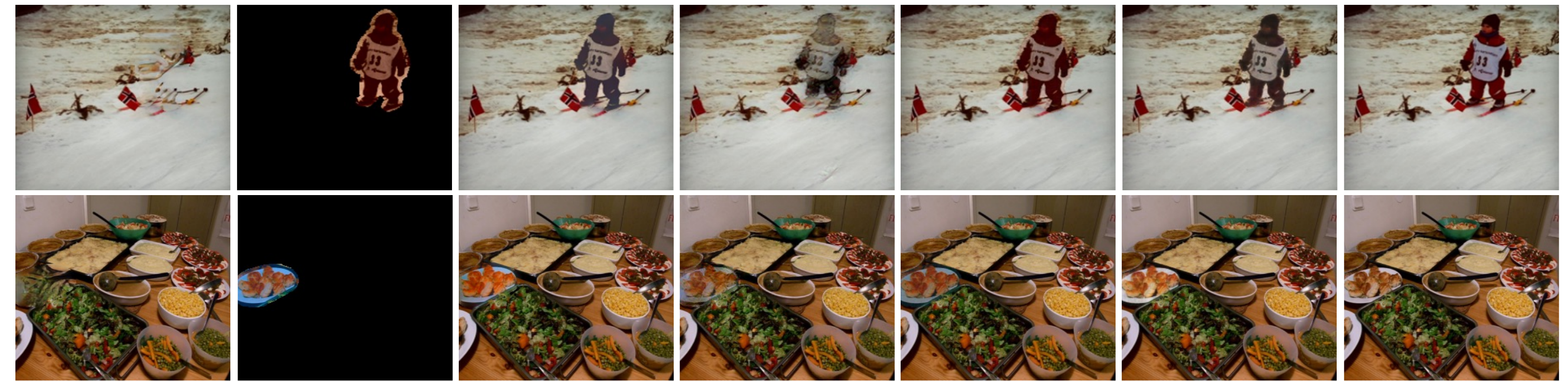}
\hspace{0.3cm}
\centering
Background\hspace{0.5cm} 
Object \hspace{0.5cm}
PB~\cite{perez2003poisson} \hspace{0.6cm} DIB~\cite{zhang2020deepblending} \hspace{0.3cm}
DoveNet~\cite{cong2020dovenet} \hspace{0.3cm}
Ours \hspace{0.3cm} Ground truth
\vspace{0.2cm}

\caption{
\textbf{Visual comparison of object composition on synthetic examples.}
We first coarsely crop the object and apply color transfer to simulate different lighting conditions \emph{(Column 2)}.
Then we compose the objects onto the \emph{same} inpainted background images shown at \emph{Column 1} (thus the ground truth images are available for evaluation). Our method produces more plausible compositions than prior approaches.
}
\label{fig:comp_gt}
\vspace{-.2cm}
\end{figure*}
 

%% file: figure_latex/05-amo-comp.tex
\begin{figure*}[t]
\begin{center}
\includegraphics[width=\linewidth]{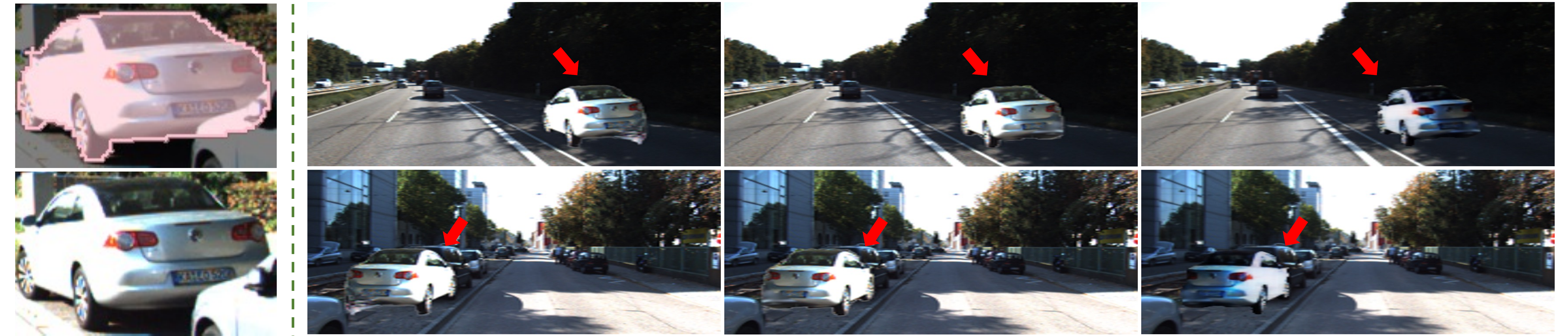} \\
\end{center}
\vspace{-.4cm}
\hspace{0.5cm} Original \hspace{1.3cm} De-occ.~\cite{zhan2020self} \hspace{1cm}
Ours (w/o composition) \hspace{1cm} Ours
\vspace{0.2cm}

\caption{
\textbf{Visual comparison of amodal instance composition.} We show the scene manipulation results for an amodal object: the original object and its ground truth amodal mask \textit{(Column 1)}, the amodal object completion and insertion by~\cite{zhan2020self} (\textit{Column 2}), ours without using the composition net $f_\text{comp}$ (\textit{Column 3}), and ours via the full method (\textit{Column 4}). 
}
\vspace{-.3cm}
\label{fig:am_harm}
\end{figure*}

%% file: 05-exp.tex
\vspace{-0.6cm}
\section{Experimental results}
\label{sec:exp}
\vspace{-0.2cm}


\topic{Datasets.}
We use the COCOA dataset~\cite{zhu2017semantic} and KINS~\cite{qi2019amodal} dataset for our evaluation. 
The COCOA dataset~\cite{zhu2017semantic} contains amodal segmentation annotations for 5,000 images from MS COCO 2014 dataset~\cite{lin2014microsoft}. 
We follow the official data split: 22,163 instances from 2,500 images for training, and 12,753 instances from 1,323 images for validation. 
The KINS dataset~\cite{qi2019amodal} was derived from KITTI~\cite{geiger2012we}, and contains 95,311 instances from 7,474 images for training, and 92,492 instances from 7,517 images for testing.

\topic{Implementation details.} 
We note that our three modules could fit with a wide range of backbone architectures. 
In practice, we adopt U-Net~\cite{ronneberger2015u} as the backbone of $f_\text{shape}$, $f_\text{comp}$, and the coarse net of $f_\text{content}$. 
Gated convolution layers ~\cite{yu2019free} and dilated convolution layers~\cite{yu2015multi} are applied in $f_\text{content}$. 
We also employ the contextual attention module from~\cite{yu2018generative} in the refine-net of $f_\text{content}$ learning to focus on object areas. 
Architecture details are presented in the supplementary.
We choose $\lambda_1=\lambda_2=1.2, \text{ and } \lambda_3=10^{-3}$ to optimize $f_\text{content}$. For $f_\text{comp}$, we use $\lambda_4=50$ for the first 150K iterations and decrease it to $5$ subsequently. 
Since instances in COCOA dataset~\cite{zhu2017semantic} is annotated in polygons, i.e., the annotations only approximate and may not present the exact object shapes, we use slightly eroded masks $m$ (with 3 pixels) after 100K iteration training to compute the first term of $\mathcal{L}_\text{mask}$ ($m$ used in the first 100K iterations). 
We also empirically choose the hyper-parameters $\lambda_5=0.005$ and $\gamma=2$. All models are trained using the two datasets at the $256 \times 256$ resolution.

\vspace{-.3cm}
\subsection{Results of amodal instance completion}
\label{sec:quant_results}
\input{figure_latex/table2-instance-completion}

\vspace{-.2cm}
\input{figure_latex/table3-composition}

\paragraph{Occluded object completion.} The object completion net, $f_\text{content}$, targets to hallucinate missing content of objects. There is no good numerical metric to evaluate content completion for the occluded object; thus, prior work~\cite{zhan2020self, ling2020variational} primarily focuses on qualitative evaluations. In an attempt to address this concern, we automate quantitative evaluations based on a property that an amodal instance completion method should faithfully reconstruct an object with part of which manually occluded. We note that the proposed evaluation strategy is commonly used in image inpainting~\cite{yu2018generative, yu2019free, yi2020contextual, yi2020contextual, nazeri2019edgeconnect}. 
Concretely, we employ three groups of representative baselines: 1) classic inpainting approaches~\cite{yu2019free,yi2020contextual}; 2) inpainting methods with auxiliary information (e.g., instance contours)~\cite{nazeri2019edgeconnect}; and 3) amodal instance completion methods~\cite{zhan2020self}. We report our evaluation in terms of the $L_1$ loss, PSNR, and SSIM. The quantitative evaluation results with $500$ test images are presented in Table~\ref{tab:content}. The numerical results indicate superior performance of our object completion module $f_\text{content}$ than the alternative~\cite{yu2019free, yi2020contextual, nazeri2019edgeconnect, zhan2020self} in this case. We show additional results and applications in the supplementary.

\vspace{-0.2cm}
\subsection{Results of amodal instance composition}
\vspace{-.1cm}
\noindent
Prior amodal completion work~\cite{zhan2020self, ling2020variational} show applications such as inserting an amodal object into a new background. 
However, they do \emph{not} consider color-inconsistency issues between the foreground object and the new background image. 
Moreover, such issues could be common as lights and shadows in the wild are exceptionally changeable. 
We proposed our composition model $f_\text{comp}$ for appearance adjustment to address this concern. 
We show a qualitative evaluation in Figure~\ref{fig:am_harm}, where the occluded white car (\textit{Column 1}) was placed in the shadow regions of the new background image.
We observe that De-occ.~\cite{zhan2020self} produced unrealistic compositing results (\textit{Column 2}) due to color inconsistency. In contrast, our entire method provided remarkable photo-realistic results (\textit{Column 4}).

As precise segmentations for the intact and the visible regions of an object may be unavailable, we expect our composition net $f_\text{comp}$ to be robust to the imperfect instances. 
For this, we verify the effectiveness of our composition module $f_\text{comp}$ on coarsely cropped COCOA validation set~\cite{zhu2017semantic} to simulate defective amodal instances. 
In this case, Poisson blending (PB)~\cite{perez2003poisson}, and DIB~\cite{ zhang2020deepblending} are employed as baseline algorithms in that they can blend inaccurately cropped objects. 
We also compare to a learning-based approach, DoveNet~\cite{cong2020dovenet}. 
One distinct difference of DoveNet~\cite{cong2020dovenet} compared to our composition module $f_\text{comp}$ lies in the format of input and output, where DoveNet~\cite{cong2020dovenet} takes as input the \emph{composite} of a background and an object image and produces a single harmonized output. 
Since DoveNet~\cite{cong2020dovenet} does not consider imprecise inputs, we fine-tuned it using the same training dataset~\cite{zhu2017semantic}. 
Figure~\ref{fig:comp_gt} presents the compositing results with the ground truth.
We observe that the blending algorithms~\cite{perez2003poisson, zhang2020deepblending} have limited performance when the compositing components have intense color contrast. 
DoveNet~\cite{cong2020dovenet} can adjust the object colors to some extent; however, it still has difficulties in dealing with coarse object boundaries. Our $f_\text{comp}$ performs stably well in terms of color and content consistency. 

We quantitatively evaluate $f_\text{comp}$ on $2,500$ pairs of coarsely segmented and color-transferred instances with their inpainted background. 
Since instance area ratios may affect the performance, we conducted the experiments on 3 disjoint ranges according to the ratio of an instance to the image, i.e., $(0.05, 0.2], (0.2, 0.4],$ and $(0.4, 0.5]$. 
We report the results in Table~\ref{tab:comp}, with variances presented in the supplementary. 
The statistics in Table~\ref{tab:comp} show that our model outperforms the baselines~\cite{sun2004poisson, zhang2020deepblending, cong2020dovenet}
on image reconstruction and photorealism. We also observe that the area ratio of an object is an influencing factor of the compositing performance, i.e., the metric scores decrease as the ratio increases. We show more qualitative results and applications in the supplementary.

\input{figure_latex/table1-amodal}

\vspace{-0.4cm}
\subsection{Results of amodal mask prediction}
\vspace{-.2cm}

We evaluate our amodal mask prediction net, $f_\text{shape}$ using the mean Intersection over Union (mIOU) metric. 
We show the quantitative results in Table~\ref{tab:shape}. 
Compared to the baselines~\cite{zhan2020self}, our amodal mask prediction net $f_\text{shape}$ performs slightly better.

\vspace{-0.2cm}
\subsection{Ablation study}
\vspace{-0.2cm}
We visualize the composition result after each of our processing step in Figure~\ref{fig:abl}. 
Concretely, we aim to insert the occluded person (the red box) to a new background. 
First, straightforward object insertion \textit{(a)} results in clearly visible artifacts.
Second, we complete the occluded regions of the person using our two modules, $f_\text{shape}$ and $f_\text{content}$, and insert the completed person to the background image \textit{(b)}. 
However, the completed object contain invalid pixels due to the imperfection of amodal mask prediction.
Third, our composition net $f_\text{comp}$ removes and harmonizes noise pixels in \textit{(b)}, achieving more plausible composition \textit{(c)}. 
Note that our composition net $f_\text{comp}$ can compose the person and the bench iteratively in a proper occlusion order.
Figure~\ref{fig:abl} suggests the necessity of our proposed pipeline. 
\input{figure_latex/ablation_fig}

We further employ an ablation study on $f_\text{comp}$ to analyze how the input and output format affect compositing performance.
In this case, we either compose the background $I_{bg}$ and the object image $I_c$ with respect to the mask or concatenate them as input, and produces either RGB or RGBA layers with the remaining architecture fixed. The results are shown in Table~\ref{tab:ablation}. 
The inferior results in the first two rows validate the necessity of the design of $f_\text{comp}$.

\vspace{-.3cm}
\subsection{Failure cases and discussions}
\input{figure_latex/07-failurecase}
While achieving favorable results than the baseline approaches, our method has several limitations. 
Specifically, the amodal instance completion task remains challenging partially due to the limited availability of training data, and complex shapes and colors of various instances. 
As shown in Figure 6, our amodal mask prediction net $f_\text{shape}$ failed to recognized the occluded body region under the table \textit{(a)}, and the object completion net $f_\text{content}$ generated unrealistic occluded content \textit{(b)} for the motorcycle. 
Second, our approach does not explicitly model environmental lighting in the wild (e.g.,  in \textit{(c)}, invalid lighting directions on hair after composition), and thus we leave it for future work.

\input{figure_latex/table4-ablation}

%% file: figure_latex/table2-instance-completion.tex
\begin{table}[t]
\centering
\footnotesize
\caption{
\textbf{\small Quantitative evaluation for amodal instance completion}. Three groups of baselines are compared:1) classic inpainting approaches~\cite{yu2019free, yi2020contextual}, 2) an inpainting method with auxiliary foreground edges~\cite{nazeri2019edgeconnect}, and 3) an amodal instance completion method~\cite{zhan2020self}.} 
\vspace{0.2cm}
\setlength{\tabcolsep}{3 pt}{
\begin{tabular}{@{}lcccccc@{}}
\toprule
\multirow{2}{*}{Method} & \multicolumn{3}{c}{COCOA~\cite{zhu2017semantic}}  & \multicolumn{3}{c}{KINS~\cite{qi2019amodal}} \\
\cmidrule(r){2-4} \cmidrule(l){5-7}
&$\ell_1$\,$\downarrow$ &PSNR$\uparrow$ & SSIM$\uparrow$  &$\ell_1$\,$\downarrow$ &PSNR$\uparrow$ & SSIM$\uparrow$  \\
\midrule
GConv~\cite{yu2019free} & 61.45  &29.40 &0.981 &47.06 &25.52 &0.935 \\
CRA~\cite{yi2020contextual} & 53.92  &30.96 &0.983 &43.82 &25.55 &0.940\\
EdgeCon.~\cite{nazeri2019edgeconnect} &41.02 &31.54 &0.983 &37.38 &26.64 &0.939 \\
De-occ.~\cite{zhan2020self} &49.58 &23.49 &0.876 & 40.72 &26.19 & 0.927\\
Ours & \textbf{37.11}  &\textbf{31.91} &\textbf{0.985} &\textbf{34.99} &\textbf{26.93} &\textbf{0.965} \\
\bottomrule
\vspace{-.2cm}
\label{tab:content}
\end{tabular}}
\vspace{-.2cm}
\end{table}

%% file: figure_latex/table3-composition.tex
\begin{table*}[t]
\caption{\textbf{Quantitative comparison for image compositing} with varying object-to-image area ratios (ranges indicated in the \emph{second row}). 
Here we compare our neural compositing model against Poisson Blending (PB)~\cite{perez2003poisson}, Deep Image Blending (DIB)~\cite{zhang2020deepblending}, and DoveNet~\cite{cong2020dovenet}. 
Note that we dilate the original object masks by 5--10 pixels and crop the objects using its dilated mask to simulate ``coarse cropping" during evaluation.
}
\vspace{.2cm}
\centering
\footnotesize
\setlength{\tabcolsep}{0.7 pt}{
\resizebox{\linewidth}{!}{
\begin{tabular}{@{}lcccccccccccccccccccccccccccc@{}}
\toprule
&\multicolumn{12}{c}{\textbf{COCOA}~\cite{zhu2017semantic}} &~
& \multicolumn{10}{c}{\textbf{KINS}~\cite{qi2019amodal}} \\
\cmidrule(r){2-12} \cmidrule(l){13-24}
Method & \multicolumn{3}{c}{(0.05, 0.2]} 
&~& \multicolumn{3}{c}{(0.2, 0.4]}  
&~& \multicolumn{3}{c}{(0.4, 0.5]}
&~& \multicolumn{3}{c}{(0.05, 0.2]} 
&~& \multicolumn{3}{c}{(0.2, 0.4]}  
&~& \multicolumn{3}{c}{(0.4, 0.5]}
\\ 
\cmidrule(r){2-4} \cmidrule(lr){6-8} \cmidrule(lr){10-12} \cmidrule(lr){14-16} \cmidrule(lr){18-20} \cmidrule(l){22-24}
&PSNR$\uparrow$ & SSIM$\uparrow$ & LPIPS$\downarrow$  
&&PSNR$\uparrow$ & SSIM$\uparrow$ & LPIPS$\downarrow$ 
&&PSNR$\uparrow$ & SSIM$\uparrow$ & LPIPS$\downarrow$ 
&&PSNR$\uparrow$ & SSIM$\uparrow$ & LPIPS$\downarrow$  
&&PSNR$\uparrow$ & SSIM$\uparrow$ & LPIPS$\downarrow$ 
&&PSNR$\uparrow$ & SSIM$\uparrow$ & LPIPS$\downarrow$\\
\midrule
PB~\cite{perez2003poisson}  
& 29.20 & 0.982 & 0.021 
&& 24.87 & 0.960  & 0.045
&& 23.21 & 0.950 & 0.058 
&& 33.07  & 0.991  & 0.012
&& 28.77 & 0.978 & 0.029 
&& 26.20 & 0.967 & 0.038 
\\
DIB~\cite{zhang2020deepblending} 
&26.80  & 0.958  & 0.049 
&&23.94  & 0.920  & 0.102  
&&21.08  & 0.881 & 0.151 
&& 29.91  & 0.978  & 0.027
&& 26.26 & 0.957 & 0.055
&& 23.76 & 0.941 & 0.069
\\
DoveNet~\cite{cong2020dovenet} 
&35.28 & 0.994  & 0.011 
&&31.62  & 0.987 & 0.022
&&30.24   & 0.983 & 0.028
&&37.47 & 0.995 & 0.007
&&36.17 & 0.994 & 0.009
&&35.52 & 0.993 & 0.010
\\	
Ours  
& \textbf{37.06}  &\textbf{0.996} &\textbf{0.008}
&&\textbf{32.87}  &\textbf{0.991} &\textbf{0.018} 
&&\textbf{31.56}  &\textbf{0.986} &\textbf{0.022} 
&&\textbf{38.60} &\textbf{0.996}  &\textbf{0.004} 	
&&\textbf{37.44}  &\textbf{0.996} &\textbf{0.007}	
&&\textbf{37.28} &\textbf{0.996} &\textbf{0.008} \\
\bottomrule
\end{tabular}}
\vspace{-0.3cm}
}
\label{tab:comp}
\end{table*}

%% file: figure_latex/table1-amodal.tex
\vspace{-0.2cm}
\begin{table}[t]
\caption{
\small \textbf{Quantitative evaluation for object amodal mask prediction using mIOU.} \textit{Raw}: no amodal mask prediction; \textit{Convex}: computing convex hull of the visible object region.}

\vspace{0.2cm}
\centering
\footnotesize
\setlength{\tabcolsep}{5 pt}{
\begin{tabular}{@{}lcccc@{}}
\toprule
Dataset  &Raw~\cite{zhan2020self} &Convex~\cite{zhan2020self} &De-occ.~\cite{zhan2020self}  & Ours \\
\midrule
COCOA~\cite{zhu2017semantic}
&0.655 &0.744 &0.814 &\textbf{0.820}\\


\bottomrule
\label{tab:shape}
\end{tabular}}
\vspace{-.6cm}
\end{table}



%% file: figure_latex/ablation_fig.tex
\begin{figure*}[h]
\small
\begin{center}
\includegraphics[width=\linewidth]{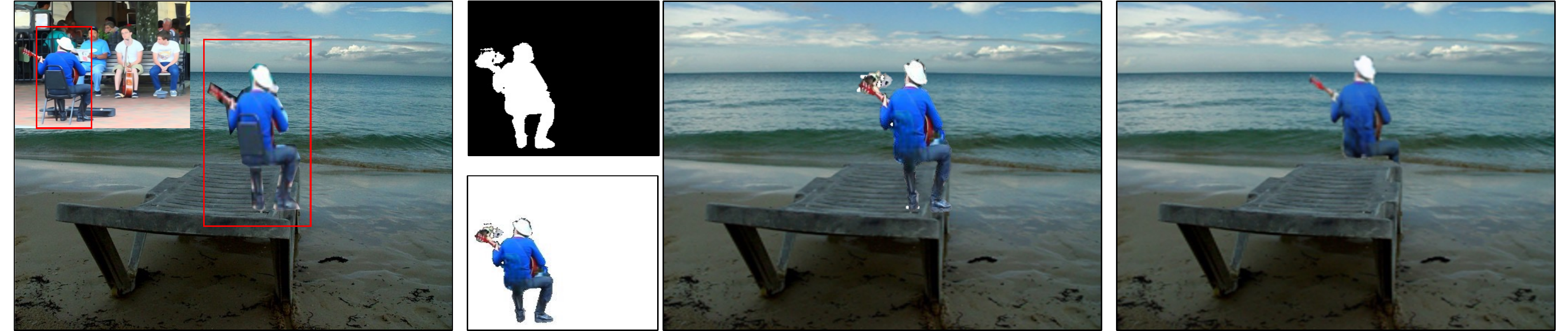} \phantom{AA}\\
\end{center}
\vspace{-.7cm}
\hspace{0.3cm} (a) Insertion (cut-n-paste)
\hspace{0.7cm} (b) Completion (w/o comp.)
\hspace{2.1cm} (c) Ours
\vspace{.2cm}
\caption{\textbf{Ablation study.} We insert the coarsely cropped occluded person (marked in the red box) to a new background via \textit{(a)} naive cut-and-paste, \textit{(b)} amodal mask prediction and object completion, and \textit{(c)} our method.}
\label{fig:abl}
\end{figure*}

%% file: figure_latex/07-failurecase.tex
\begin{figure}[t]
\small
\begin{center}
\includegraphics[width=\linewidth]{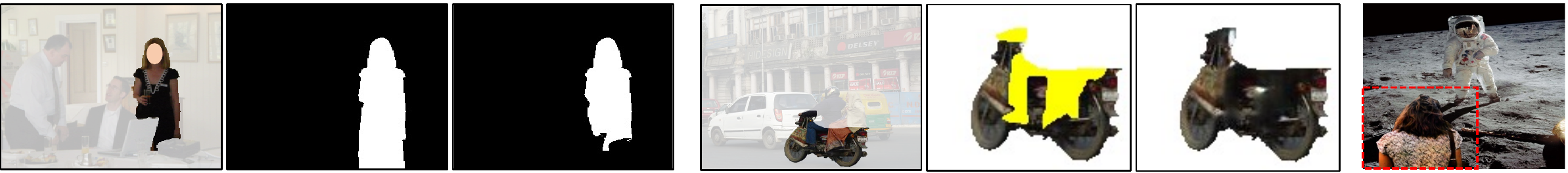}\phantom{AA}\\
\end{center}
\vspace{-.3cm}
\hspace{0.2cm} \textit{(a)} Object \hspace{.8cm} GT \hspace{1.2cm} Ours
\hspace{.8cm} \textit{(b)} Object \hspace{.3cm} Occ. region \hspace{.5cm} Ours
\hspace{.4cm} \textit{(c)} Comp. result
\vspace{.25cm}
\caption{\textbf{Failure cases} of our amodal mask prediction net $f_\text{shape}$ \textit{(a)}, the object completion net $f_\text{content}$ \textit{(b)}, and the neural composition net $f_\text{comp}$ \textit{(c)}.
}
\label{fig:failure}
\end{figure}

%% file: figure_latex/table4-ablation.tex
\begin{table}[h]
\caption{\textbf{Ablation study} in the image composition task using 500 KINS test images~\cite{qi2019amodal}. The input is either the \emph{composition} of the background image and the object image with respect to the mask, or the \emph{concatenation} of the two.}
\vspace{.2cm}
\centering
\footnotesize
\begin{tabular}{@{}llccc@{}}
\toprule
Input & Output & PSNR $\uparrow$ & SSIM $\uparrow$ & LPIPS $\downarrow$  \\
\midrule
Composition & RGB  &34.56 &0.992 &0.012 \\
Composition & RGBA &36.23 &0.993 &0.011 \\
Concatenation & RGBA &\textbf{37.36} &\textbf{0.996} &\textbf{0.007} \\
\bottomrule
\end{tabular}
\vspace{-.5cm}
\label{tab:ablation}
\end{table}

%% file: 06-con.tex
\vspace{-.3cm}
\section{Conclusions}
\label{sec:con}
\vspace{-.2cm}
\noindent
We proposed a fully automatic system for image composition that is capable of handling imperfect and heterogeneous amodal inputs. 
Experimental results demonstrate that our approach outperforms various baselines for dealing with imperfect and heterogeneous amodal instances. 
From an application perspective, our method has a broad impact on amodal instance manipulation and style harmonization. Beyond that, we leave automatic adjustments for spatial transformations and better handling of lighting for future work.

%% file: 07-sup.tex
\subsection*{Supplementary}

In this supplementary document, we provide the network architectures and additional implementation details to complement the main paper. 
We also present additional quantitative and qualitative comparisons.
We will make the source code and pretrained models publicly available to foster future research.

\subsection*{A.Implementation details}

\subsection*{A.1 Network architecture}
We adopt U-Net~\cite{ronneberger2015u} as the backbone of the amodal mask prediction net $f_\text{shape}$ and the composition net $f_\text{comp}$. 
We show the full backbone in Table~\ref{tab:unet}.
We use ``zero-padding", batch normalization (``bn"), convolutional transpose (``convt"), and skip connection (``skipk" refers to a skip connection with layer k) in the neural compositing network.
The full definition of the neural compositing network is defined as follows.

We modify the CRA model~\cite{yi2020contextual} as the backbone of our object content completion net $f_\text{content}$. 
We note that CRA's architecture is a common paradigm in image inpainting tasks, as similar structures were widely used in ~\cite{yu2019free, xiong2019foreground, zeng2020high}. 
There are, however, several important differences between our object content completion net $f_\text{content}$ and the CRA model ~\cite{yi2020contextual}. 
First, the goal of the two methods is different: our object content completion net $f_\text{content}$ aims to hallucinate the \emph{invisible regions} of an object from the visible regions
On the other hand, the CRA method~\cite{yi2020contextual} targets to recover the entire image (i.e., filling in the missing regions using all the remaining known pixels as contexts). 
Second, the inputs of the two methods are different. 
Specifically, the input of our object content completion module, $f_\text{content}$, is a triplet consisting of a masked object image ($I_o \odot m_{vis}$ in the main manuscript), and the two binary masks ($m_\text{vis}$ and $m_\text{inv}$ in the main manuscript). 
The CRA model~\cite{yi2020contextual} takes as input a masked image as well as the corresponding mask. 
Consequently, training data pre-processing for the object completion net $f_\text{content}$ in our method is different from ~\cite{yi2020contextual}. 
In our work, we randomly hide part of the target object to simulate amodal instance completion. 
In particular, we randomly sample an object masks $m_\text{inv}$ from the dataset and use them to occlude part of the target object $I_o$.
We apply basic transformations, e.g., scaling and translation, to the sampled object mask $m_\text{inv}$ such that it has overlap with the target object $I_o$. 
Table 2 and Figure 3 in the main manuscript show effectiveness of the modifications on the object completion model $f_\text{content}$ compared to the original CRA~\cite{yi2020contextual}. We use ``same" padding and the Exponential Linear Unit (ELU) activation function~\cite{clevert2015fast} for all convolution layers. 
We show the full definition of the object content completion net $f_\text{content}$ in Table~\ref{tab:coarse}-\ref{tab:refine}.

\begin{table}
\caption{The backbone used in the amodal shape prediction net $f_\text{shape}$ and the neural composition net $f_\text{comp}$. 
\textit{bn}: batch normalization, \textit{convt}: convolutional transpose, \textit{skipk}: a skip connection with layer k.}
\label{tab:unet}
\vspace{.2cm}
\centering
\begin{tabular}{cccc} 
 \toprule
layers & out channels & stride & activation\\
 \midrule
  $4\times 4$ conv & 64 & 2 & leaky\\
 $4\times 4$ conv, bn & 128 & 2 & leaky\\
 $4\times 4$ conv, bn & 256 & 2 & leaky\\
 $4\times 4$ conv, bn & 256 & 2 & leaky\\
 $4\times 4$ conv, bn & 256 & 2 & leaky\\
 $4\times 4$ conv, bn & 256 & 1 & leaky\\
 $4\times 4$ conv, bn & 256 & 1 & leaky\\
 skip5, $4\times 4$ convt, bn & 256 & 2 & relu\\
 skip4, $4\times 4$ convt, bn & 256 & 2 & relu\\
 skip3, $4\times 4$ convt, bn & 128 & 2 & relu\\
 skip2, $4\times 4$ convt, bn & 64 & 2 & relu\\
 skip1, $4\times 4$ convt, bn & 64 & 2 & relu\\ 
 $4\times 4$ conv & 4 & 1 & tanh\\
 \bottomrule
\end{tabular}
\end{table}

\begin{table}
\caption{The \textit{coarse network} of the object content completion module $f_\text{content}$. \textit{num} refers to the number of layers. \textit{out channels} refers to the number of output channels after the layer. \textit{stride} and \textit{dilation} are the parameters of the convolution operation. }
\label{tab:coarse}
\footnotesize
\vspace{.2cm}

\centering 
\begin{tabular}{cccccc} 
 \toprule
layers &num & out channels & stride &dilation &out shape \\
 \midrule
$5\times 5$ gconv & 1 & 32 &2 & 1 & $128\times 128$ \\
$3\times 3$ gconv & 1 & 64 &1 & 1 & $128\times 128$ \\
$3\times 3$ gconv & 1 & 64 &2 & 1 & $64\times 64$ \\
$3\times 3$ gconv & 6 & 64 &1 & 1 & $64\times 64$ \\
$3\times 3$ gconv & 5 & 64 &1 & 2 & $64\times 64$ \\
$3\times 3$ gconv & 4 & 64 &1 & 4 & $64\times 64$ \\
$3\times 3$ gconv & 2 & 64 &1 & 8 & $64\times 64$ \\
$3\times 3$ gconv & 3 & 64 &1 & 1 & $64\times 64$ \\
$3\times 3$ deconv & 1 & 32 &1 & 1 & $128\times 128$ \\
$3\times 3$ deconv & 1 & 3 &1 & 1 & $256 \times 256$\\
\bottomrule
\end{tabular}
\end{table}

\begin{table}
\caption{The \textit{refine network} of the object content completion module $f_\text{content}$. The notations in the first row are identical to Table~\ref{tab:coarse}. $attn$ refers to Contextual Attention~\cite{yu2019free}.}
\label{tab:refine}
\footnotesize
\vspace{.2cm}
\centering
\begin{tabular}{cccccc} 
 \toprule
layers &num & out channels & stride &dilation &out shape \\
 \midrule
$5\times 5$ gconv & 1 & 32 &2 & 1 & $128\times 128$ \\
$3\times 3$ gconv & 1 & 32 &1 & 1 & $128\times 128$ \\
$3\times 3$ gconv & 1 & 64 &2 & 1 & $64\times 64$ \\
$3\times 3$ gconv & 1 & 128 &2 & 1 & $32\times 32$ \\
$3\times 3$ gconv & 2 & 128 &1 & 1 & $32\times 32$ \\
$3\times 3$ gconv & 1 & 128 &1 & 2 & $32\times 32$ \\
$3\times 3$ gconv & 1 & 128 &1 & 4 & $32\times 32$ \\
$3\times 3$ gconv & 1 & 128 &1 & 8 & $32\times 32$ \\
$3\times 3$ gconv & 1 & 128 &1 & 16 & $32\times 32$ \\
$3\times 3$ gconv + attn & 1 & 128 &1 & 1 & $32\times 32$ \\
$3\times 3$ deconv & 1 & 64 &1 & 1 & $64\times 64$ \\
$3\times 3$ gconv + attn & 1 & 64 &1 & 1 & $64\times 64$ \\
$3\times 3$ deconv & 1 & 32 &1 & 1 & $128\times 128$ \\
$3\times 3$ gconv + attn & 1 & 32 &1 & 1 & $128\times 128$ \\
$3\times 3$ deconv & 1 & 3 &1 & 1 & $256\times 256$ \\
\bottomrule
\end{tabular}
\end{table}

\subsection*{A.2 Training details}
We use the Adam optimizer~\cite{kingma2014adam} to train the three networks. 
The initial learning rate is 1$e$-3 for the amodal prediction net $f_\text{shape}$, 1$e$-4 for the object content completion model $f_\text{content}$, and 2$e$-4 for the neural composition net $f_\text{comp}$. 
We select $\text{beta1}=0.5$ and $\text{beta2}=0.9$ for all Adam optimizer~\cite{kingma2014adam}. 

\subsection*{A.3 Inference algorithm}
We present the inference procedure in Algorithm ~\ref{alg:1}.

\begin{algorithm}
	\caption{Inference Procedure}
	\label{alg:1}
	\begin{algorithmic}[1]
		\Input{
		an amodal mask prediction model $f_{\text{shape}}$; an object completion model, denoted as $f_{\text{content}}$; a neural compositing model, denoted as $f_{\text{comp}}$; input background $I_{bg}$ and an ordered collection of object images $X = \{x^{(1)}, x^{(2)}, ..., x^{(M)}\}$, where $x^{(j)} = (I^{(j)}, m_{vis}^{(j)})$, $j \in \{1, 2,...,M\}$ and $M= |X|$ (note that $m_{vis}^{(j)}$ is not required to be precise during inference).
		}
		\For{$j = 1, \dots, M$}
		   \If{$x^{(j)}$ is occluded}
    		    \State{$\hat{m}_{vis}, \hat{m}_{amodal} = f_\text{shape}(x_j, m_{vis})$}
    		    \State{$\hat{I}_{o} = f_\text{content}(I, \hat{m}_{vis}, \hat{m}_{amodal})$}
		   \Else
		        \State $\hat{I}_{o} = I \odot m_{vis}$
		 \EndIf
		   \State{$I_{out}, \alpha = f_\text{comp}(I_{bg}, \hat{I}_{o})$} 
		   \State{$\hat{I} =  \alpha \odot I_{out} + (1 -  \alpha) \odot I_{bg}$} 
        \EndFor
        \RETURN {$\hat{I}$} 
	\end{algorithmic}  
\end{algorithm}

\subsection*{B. Ablation study}
\begin{figure}
    \centering
    \includegraphics[width=\linewidth]{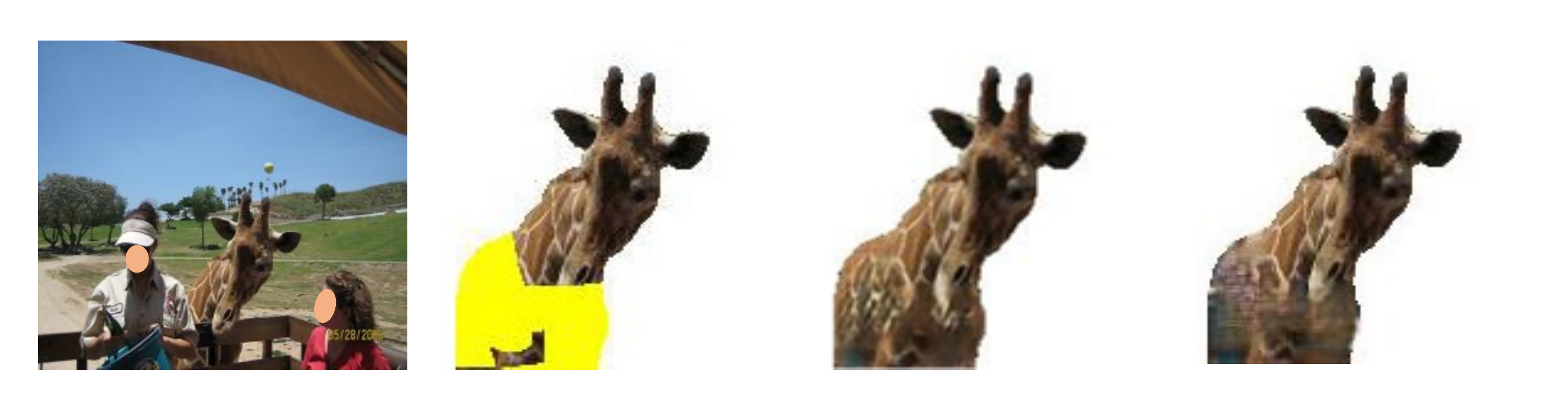}
    \begin{minipage}{1\linewidth}
    \hspace{1cm} Original \hspace{1cm} Occ. object \hspace{1.5cm} \textbf{Ours w/o bg} \hspace{1.5cm} Ours w/ bg $\qquad$
    \end{minipage}
    \vspace{.1cm}
    \caption{\textbf{Ablation study of the object completion model $f_\text{content}$.} We predict occluded regions of the objects. \textit{Column 1}: Original images from COCOA validation dataset~\cite{zhu2017semantic} with visible and amodal masks annotated by ~\cite{zhu2017semantic}; \textit{Column 2}: objects with occluded region marked in yellow and background region in white; \textit{Column 3}: the object completion model with $I \odot  m_{vis}$ as input (i.e., background pixels are empty); \textit{Column 4}: the object completion model with $I \odot (1 - m_{inv})$ as input (i.e., background pixels are valid). }
    \label{fig:abl}
\end{figure}

We conduct an ablation study for the content completion model $f_\text{content}$ where the new inputs are triplet images consisting of a masked image with the background region preserved ($I \odot m_{inv}$), a visible mask ($m_{vis}$), and an invisible mask ($m_{inv}$). 
In other words, we do \emph{not} hide the background region during training. 
We train the new content completion model with an identical number of iterations as the prior model.  
We show the comparison results in Figure~\ref{fig:abl} which indicates that \textbf{removing background regions benefits object reconstruction}.

\begin{figure}
 \vspace{-.2cm}
    \centering
    \includegraphics[width=\linewidth]{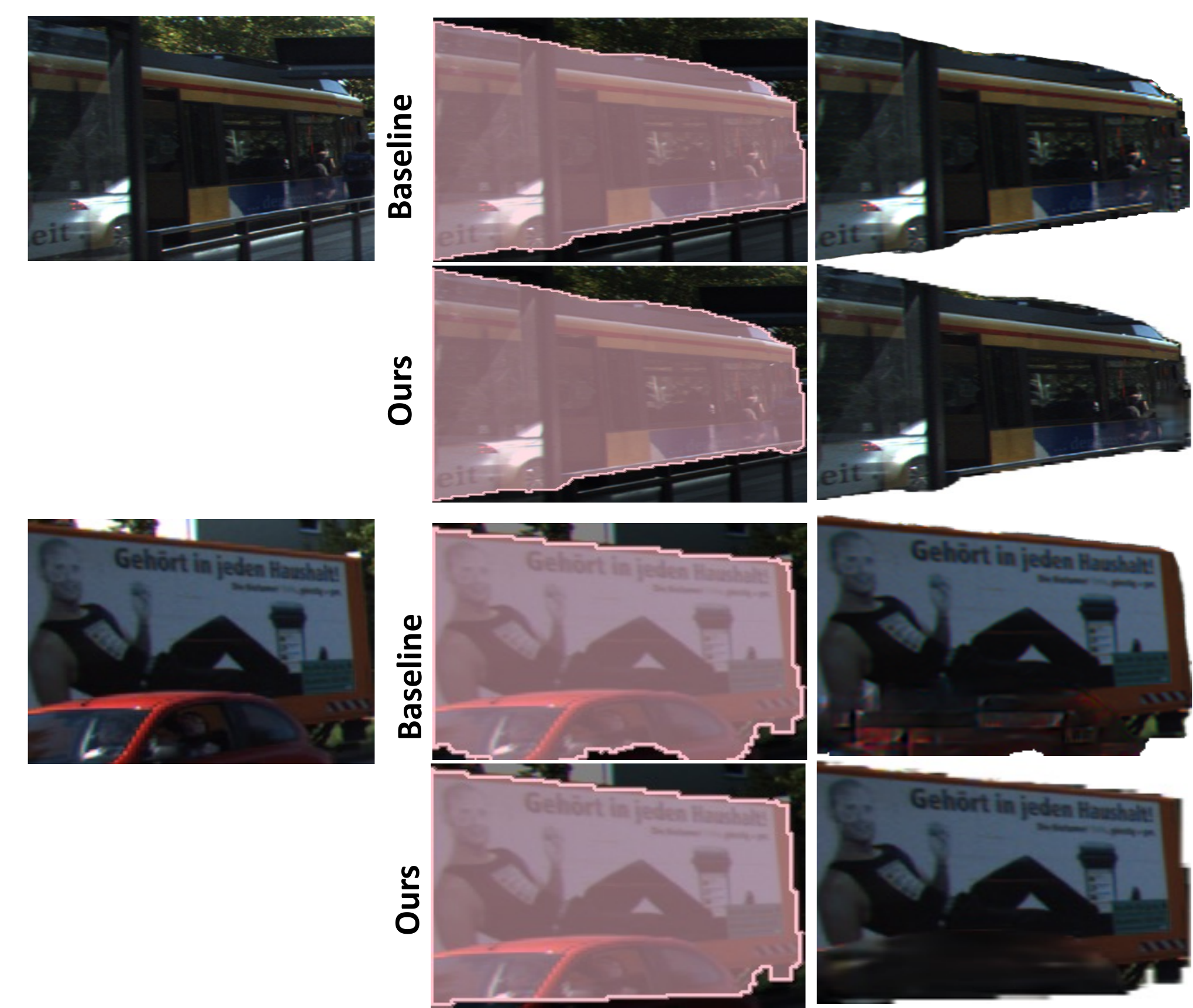}
    
    \vspace{.2cm}
    \caption{\textbf{Results of amodal instance completion on KINS test dataset.} \textit{Column 1}: Original objects, \textit{Column 2}: predicted amodal masks, \textit{Column 3}: synthetic amodal objects.}
    \label{fig:kins_obj}
\end{figure}

\subsection*{C. Additional results}

\subsection*{C.1 Variance of the comparison for image composition.}
We computed Table 3 in the main manuscript by repeating the experiments three times with the object styles transferred towards a randomly selected reference image. 
Here, we show the variances of the statistical results in Table~\ref{tab:comp_var}.

\begin{table}[h]
\caption{\textbf{Variances of the quantitative comparison for image compositing on the COCOA validation dataset~\cite{lin2014microsoft}.} Three baselines are compared: PB~\cite{perez2003poisson}, DIB~\cite{zhang2020deepblending}, and DovNet~\cite{cong2020dovenet}.
}
\vspace{.2cm}
\centering
\small
\resizebox{\linewidth}{!}{
\begin{tabular}{lccccccccccccc}

\toprule
& \multicolumn{3}{c}{(0.05, 0.2]} 
&~& \multicolumn{3}{c}{(0.2, 0.4]}  
&~& \multicolumn{3}{c}{(0.4, 0.5]}
\\ 
\cmidrule{2-4} \cmidrule{6-8} \cmidrule{10-12}
 &PSNR $\uparrow$ & SSIM $\uparrow$ & LPIPS $\downarrow$
&&PSNR $\uparrow$ & SSIM $\uparrow$ & LPIPS $\downarrow$
&&PSNR $\uparrow$ & SSIM $\uparrow$ & LPIPS $\downarrow$ \\
\midrule

~\cite{perez2003poisson}  
&1$e$-2  &2$e$-7  &2$e$-7  
&&5$e$-2  &2$e$-7  &9$e$-7  
&&1$e$-1  &4$e$-6  &3$e$-6   \\
~\cite{zhang2020deepblending}  
&6$e$-2   &3$e$-6   &3$e$-4   
&&2$e$-1    &3$e$-5   &3$e$-4   
&&3$e$-1    &2$e$-4   &2$e$-5 \\
~\cite{cong2020dovenet}  
&3$e$-3  &3$e$-7  &2$e$-8   
&&2$e$-2  &4$e$-8. &5$e$-7   
&&2$e$-2 &1$e$-7 &2$e$-6 \\
Ours  
&{7$e$-2}  &{2$e$-7} &{4$e$-7}
&&1$e$-2  &2$e$-8 &8$e$-8
&&8$e$-2  &1$e$-6 &2$e$-6 \\
\bottomrule
\end{tabular}
}
\label{tab:comp_var}
\end{table}

\subsection*{C.2 Amodal instance composition on KINS dataset}
\begin{figure}
    \centering
    \includegraphics[width=0.95\linewidth]{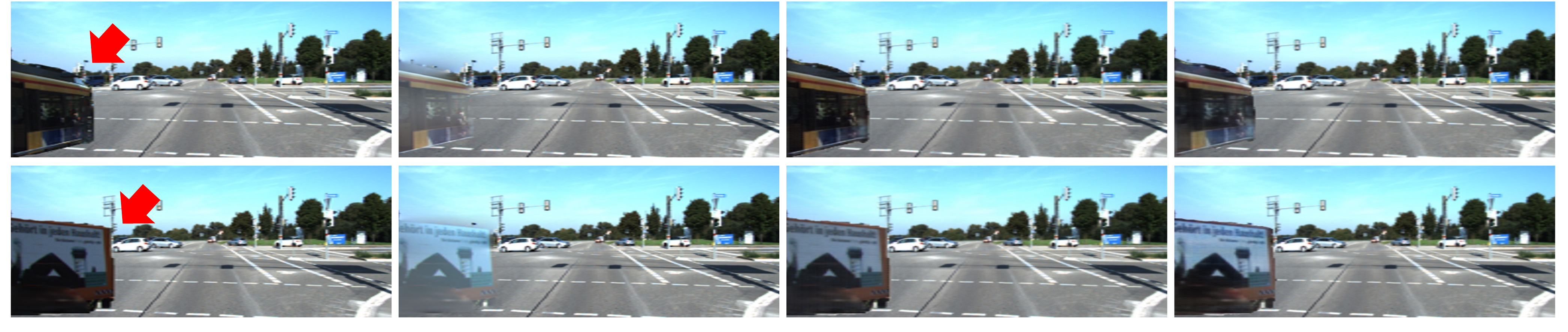}
    
    \begin{minipage}{\linewidth}
     \hspace{.7cm}De-occ.~\cite{zhan2020self} \hspace{1.5cm} PB~\cite{perez2003poisson}\hspace{2cm} DoveNet~\cite{cong2020dovenet} \hspace{2cm} Ours
    \end{minipage}
    \vspace{.2cm}
    
    \caption{\textbf{Results of amodal instance composition on KINS test dataset.} We inserted the instances into the new background image (marked by red arrow). \textit{De-occ.}~\cite{zhan2020self} generated the intact object and then directly pasted the amodal instance into the new background. \textit{PB} and \textit{DoveNet}~\cite{perez2003poisson,cong2020dovenet} were used to harmonize into the new background the amodal instance by our amodal mask prediction net $f_\text{shape}$ and object content completion net $f_\text{content}$. \textit{Ours} were achieved by the proposed three modules. }
    \label{fig:kins_comp}
\end{figure}

We show additional amodal mask prediction and object content completion comparisons in Figure~\ref{fig:kins_obj} and composition results of the identical instances in Figure~\ref{fig:kins_comp}.
Specifically, in Figure~\ref{fig:kins_obj}, we present the predicted amodal masks and the hallucinated results by the baseline method~\cite{zhan2020self} and our approach in column 2-3. 
Our amodal mask prediction net $f_\text{shape}$ has a comparable performance with~\cite{zhan2020self}, and our content completion net $f_\text{content}$ can hallucinate the occluded regions of the objects with fewer artifacts (see the occluded corners and the bottom of the vehicles). 
In Figure~\ref{fig:kins_comp}, we insert the completed objects into a new background image (marked by the red arrow). De-occ.~\cite{zhan2020self} does not harmonize the amodal instances with the new background in its applications, leading to unrealistic compositing results. 
We also notice that PB~\cite{perez2003poisson} performs unsatisfied when the objects have obvious color contrast with the background. 
The same issues of PB~\cite{perez2003poisson} are discussed in prior work~\cite{zhang2020deepblending, tao2013error}. 
In contrast, our composition net $f_\text{comp}$ works stably well with fewer artifacts.

\subsection*{C.3 Additional applications}

Here we demonstrate several additional applications using our method.

\begin{figure}
\includegraphics[width=0.98\linewidth]{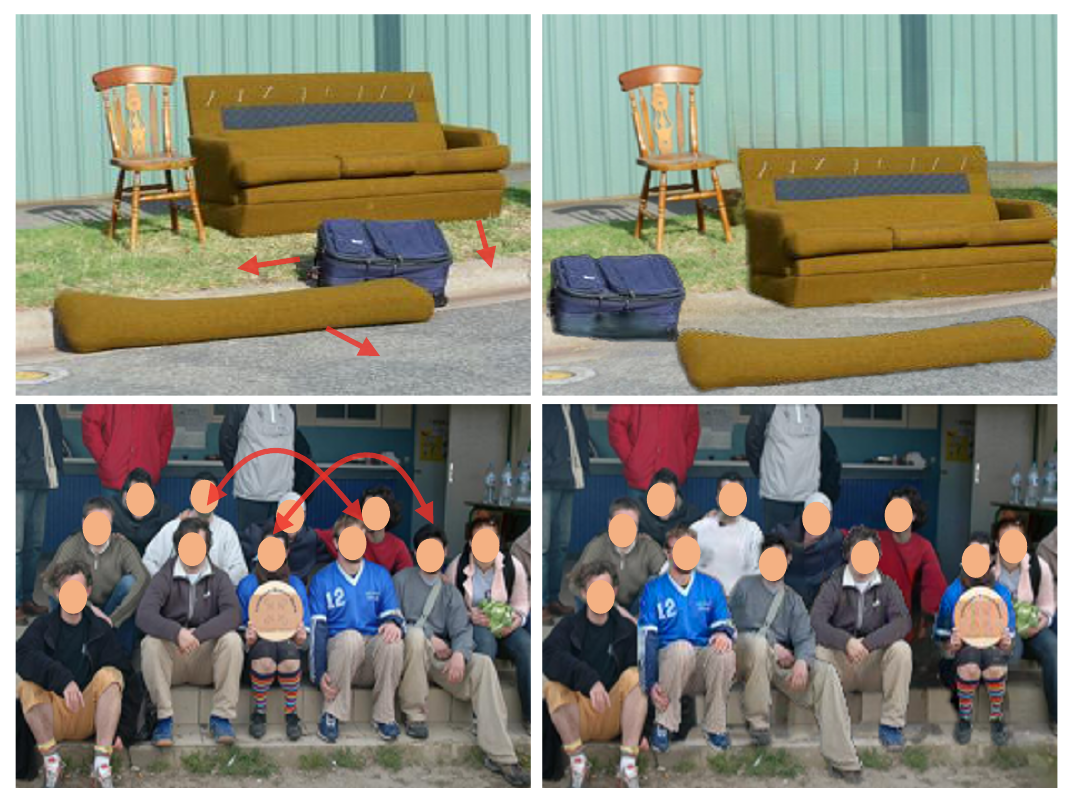} 

\vspace{0.2cm}
\caption{\textbf{Object re-shuffling results}. \textit{left}: original images. The arrows indicate object moving directions; \textit{right}: results with objects re-shuffled.}
\label{fig:obj_reshuf}
\end{figure}

\topic{Object re-shuffling.} 
Our object completion model enables re-shuffling objects. Figure~\ref{fig:obj_reshuf} shows examples of re-shuffling objects to new locations in the images. 
A limitation of our model is that the occluded region is smooth, e.g., the blue luggage in Figure~\ref{fig:obj_reshuf}. 
We leave this for future improvement.

\begin{figure}[h]
\includegraphics[width=0.98\linewidth]{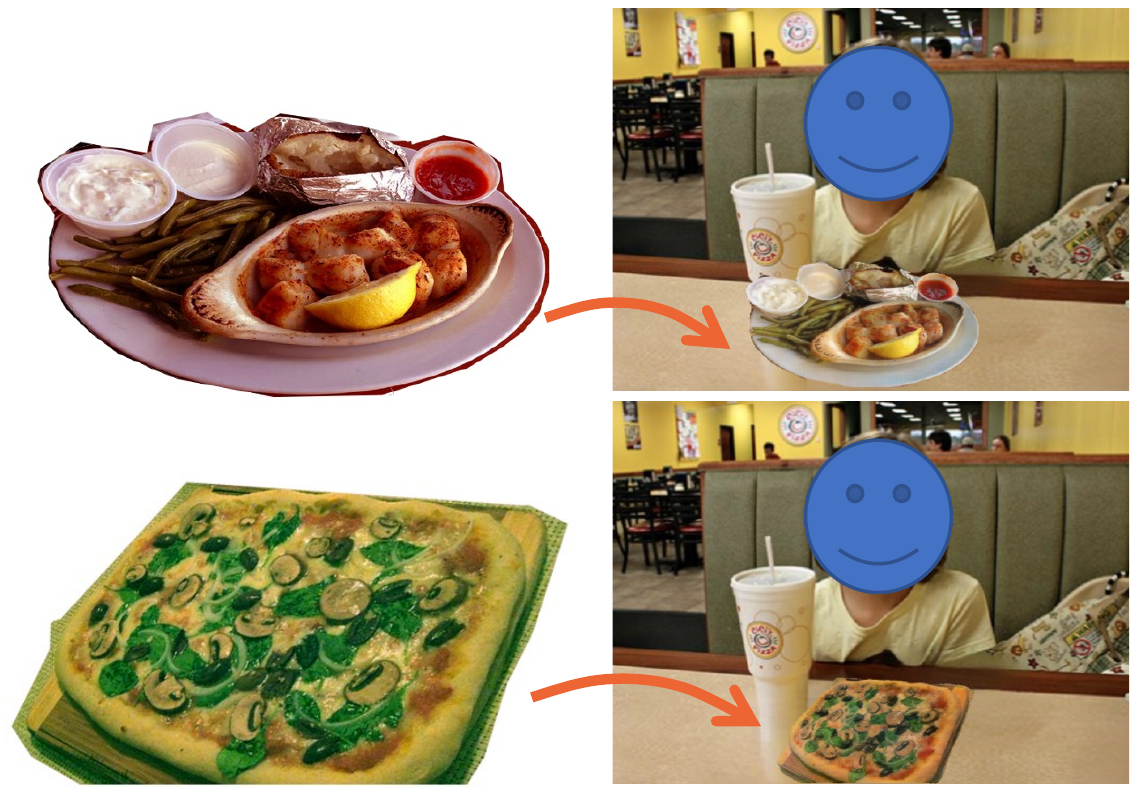}
\vspace{0.2cm}
\caption{\textbf{Object insertion results}. 
We insert the dishes into the background image. Our method harmonizes the content and refines the imperfect masks.
}

\label{fig:obj_insersion2}
\end{figure}

\topic{Object insertion with imperfect inputs.}
In Figure~\ref{fig:obj_insersion2}, we show the results of placing two dishes onto an indoor scene. 
Our method automatically adjusts the foreground instances' colors towards the background images with redundant pixels around objects removed in these cases.

